\documentclass[journal]{IEEEtran}
\usepackage{cite}
\ifCLASSINFOpdf
\else
\fi

\usepackage{array}
\usepackage[caption=false,font=footnotesize]{subfig}

\usepackage{times}
\usepackage{epsfig}
\usepackage{graphicx}
\usepackage{amsmath}
\usepackage{amssymb}
\usepackage{epstopdf}
\usepackage{multirow}
\usepackage{CJK}

\usepackage{subfig}
\usepackage{color}

\begin{document}
%
\title{A Central Difference Graph Convolutional Operator for Skeleton-Based Action Recognition}
%
%
%

\author{Shuangyan Miao,
        Yonghong Hou,
        Zhimin Gao,
        Mingliang Xu,
        and Wanqing Li
\thanks{Manuscript received XX XX, 2021; revised XX XX, 2021. Corresponding author: Zhimin Gao}
\thanks{S. Miao and Y. Hou are with School of Electrical and Information Engineering, Tianjin University, Tianjin, China (e-mail: symiao@tju.edu.cn; houroy@tju.edu.cn).}
\thanks{Z. Gao and M. Xu are with School of Information Engineering, Zhengzhou University, Zhengzhou, China (e-mail: iegaozhimin@zzu.edu.cn; iexumingliang@zzu.edu.cn).}
\thanks{W. Li is with Advanced Multimedia Research Lab, University of Wollongong, Wollongong, Australia (e-mail: wanqing@uow.edu.au).}}

%
%

\markboth{IEEE TRANSACTIONS ON CIRCUITS AND SYSTEMS FOR VIDEO TECHNOLOGY,~Vol.~X, No.~X, X~2021 }%
{Shell \MakeLowercase{\textit{et al.}}:A Central Difference Graph Convolutional Network for Skeleton-Based Action Recognition}
%



\maketitle

\IEEEpubid{\begin{minipage}{\textwidth}\ \\ \\ \\ \\[8pt] \centering
Copyright $\copyright$ 20xx IEEE. Personal use of this material is permitted. However, permission to use this material for any other purposes must be obtained from the IEEE by sending an email to pubs-permissions@ieee.org.
\end{minipage}}

\begin{abstract}

This paper proposes a new graph convolutional operator called central difference graph convolution (CDGC) for skeleton based action recognition. It is not only able to aggregate node information like a vanilla graph convolutional operation but also gradient information. Without introducing any additional parameters, CDGC can replace vanilla graph convolution in any existing Graph Convolutional Networks (GCNs). In addition, an accelerated version of the CDGC is developed which greatly improves the speed of training. Experiments on two popular large-scale datasets NTU RGB+D 60 \& 120 have demonstrated the efficacy of the proposed CDGC. Code is available at https://github.com/iesymiao/CD-GCN.
\end{abstract}


%
\IEEEpeerreviewmaketitle

\section{Introduction}
%
%
%
%

\IEEEPARstart{H}{uman} action recognition is a pupular and challenging research topic. It has attracted extensive attention in different research fields in recent years~\cite{popoola2012video,ramanathan2014human,wang2018rgb,zhang2016rgb,ji2019survey} due to its wide range of applications such as video analytic, robotics, human-machine interactions and health and aged care. Skeleton is one of the effective representations of human motion. It is not only robust against interference from background but also has computational advantages, as each skeleton consists of a small number of joints.  Therefore, much recent research is based on 3D skeletons~\cite{wang2016action,hou2016skeleton,li2017joint,cao2018skeleton,li2017skeleton,tang2018online,li2018multiview,jiang2019action}.

\begin{figure*}[!t]
	\begin{center}
		\includegraphics[height = 42mm, width = 152mm]{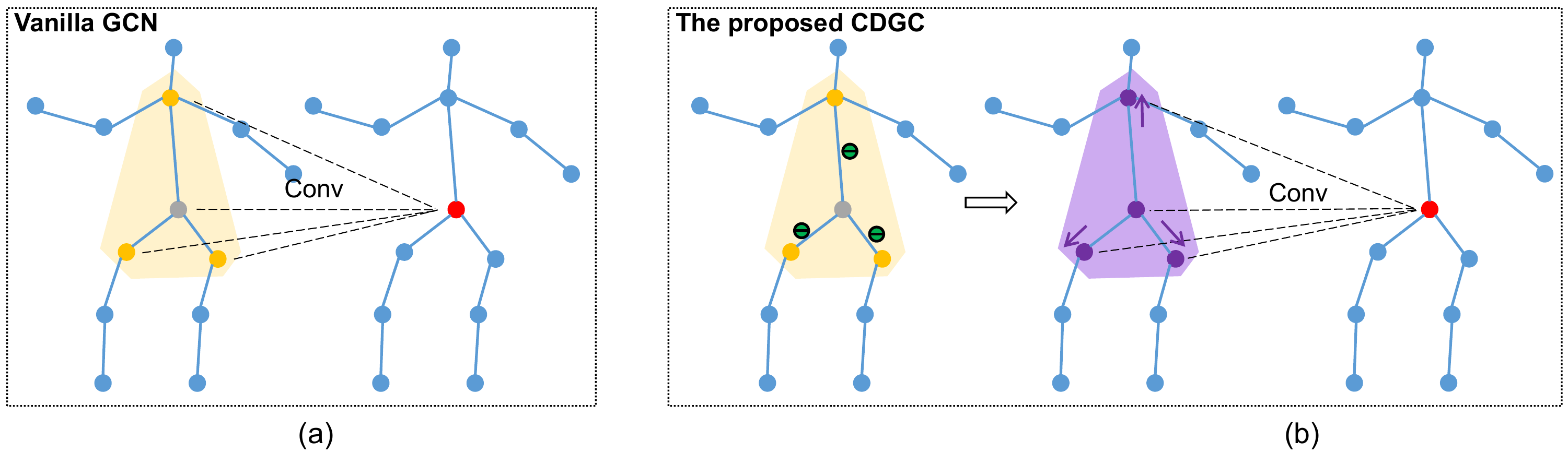}
	\end{center}
	\setlength{\abovecaptionskip}{-0.2cm}
	\caption{Illustration of feature aggregation in a vanilla GCN and proposed CDGC. \textbf{(a)} Vanilla GCN. The yellow nodes in the yellow sampling area are the adjacent nodes of the central node (grey node). The red node is the one whose feature is updated through convolution. \textbf{(b)} The proposed CDGC. The green minus signs represent taking differences between the features of the adjacent nodes and the central node. The purple nodes represent their gradient features relative to the central node and arrows indicate the corresponding gradient directions.}
	
\label{CDGC}
\end{figure*}

Among the reported approaches to skeleton based action recognition, graph convolutional network (GCN) has shown much more promising performance than convolutional neural network (CNN) and/or recurrent neural network (RNN) based approaches~\cite{li2019actional,yan2018spatial,cheng2020skeleton,cao2018skeleton,jiang2019action}, mainly due to the fact that a skeleton can be naturally considered as a graph where each joint being a node and an anatomic link between two joints being an edge in the graph. To date, many GCN-based methods take joint locations or coordinates directly as input of their models. In~\cite{shi2019two}, bones calculated from adjacent and anatomically connected joints are introduced as an additional input modality for the first time, leading to an improvement of the recognition performance. However, both joints and bones are defined by body's anatomic structure. On one hand, extensive research has been conducted on improving the topology of a skeleton graph by including dynamic links between joints. For example, 2s-AGCN~\cite{shi2019two} employs additional non-local neural network modules to learn an effective graph topology. AS-GCN~\cite{li2019actional} proposes an encoder-decoder structure to capture action-specific links. On the other hand, many works adopting GCN attempt to extract richer features by either expanding the space-time perception domain~\cite{wu2021graph2net} or devising complex multi-stream network architectures~\cite{song2020richly,kong2021symmetrical}. However, these methods directly use vanilla graph convolution operations to aggregate information from the associated nodes in the graph topology for the central node, ignoring the local motion information between the central node and the neighboring nodes, as shown in Fig. \ref{CDGC}(a).

Our hypothesis is that improving the graph convolution operator itself via incorporating the local motion information among the nodes could be as effective as improving the topology of a skeleton graph. To this end, this letter proposes a novel graph convolution operator called Central Difference Graph Convolution (CDGC) for skeleton based action recognition. It can capture the dynamic gradient information related to the local motion between the central node and its associated neighboring nodes during feature aggregation, as shown in Fig. \ref{CDGC}(b). Besides, the proposed CDGC does not introduce any additional parameters and it can be used to replace vanilla graph convolution in any existing graph convolution neural networks to improve their performance. In addition, an accelerated version of CDGC is developed, which greatly improves the training speed of the network. Extensive experiments on two large-scale benchmark datasets have demonstrated the efficacy of the proposed CDGC.

\section{Proposed Method}
\label{proposed method}
This section first gives an overview of the Vanilla Graph Convolution in Sec. \ref{Vanilla Graph Convolution}. The proposed Central Difference Graph Convolution (CDGC) and its implementation are detailed in Sec. \ref{Central Difference Graph Convolution} and \ref{Implementation}, respectively. Its accelerated version is presented in Sec. \ref{Accelerated Version of CDGC}.

\subsection{Vanilla Graph Convolution}
\label{Vanilla Graph Convolution}

A graph is usually represented as $G=(V, E)$, where $V=\left\{v_{1}, \ldots, v_{N}\right\}$ is a set of $N$ nodes representing joints of a skeleton and $E$ is a set of edges representing bones between joints. Given a skeleton graph, multiple layers of vanilla graph convolutions are applied to extract multi-level features. Suppose $v_{i}$ is a vertex in the graph, a vanilla graph convolution operation centred at it is expressed as:
\begin{equation}
\textbf{y}\left(v_{i}\right)=\sum_{v_{j} \in R_{i}} \frac{1}{Z_{i j}} \textbf{x}\left(v_{j}\right) \cdot \textbf{w}\left(l_{i}\left(v_{j}\right)\right)
\label{equation1}
\end{equation}
where $\textbf{x}$ and $\textbf{y}$ denote an input feature map and output feature map, respectively. $\textbf{w}$ is the weight function that provides a weight vector based on the given input. $R_{i}$ is the 1-distance neighboring vertexes of $v_{i}$. $l_{i}$ is a partitioning function to partition vertexes in $R_{i}$ into a fixed number of subsets, with each subset having its own convolution weight vectors. In the classic partition strategy \cite{yan2018spatial}, $R_{i}$ is divided into three subsets: $S_{1}$, $S_{2}$, and $S_{3}$. In specific, $S_{1}$ is $v_{i}$ itself; $S_{2}$ is the centripetal subset which contains vertexes closer to skeleton gravity center in $R_{i}$; and $S_{3}$ is the centrifugal subset which contains vertexes away from skeleton gravity center. $Z_{i j}$ denotes a normalizing term which balances the contribution of each subset to the output.

In general, there are two main steps in the vanilla graph convolution: 1) sampling neighboring vertexes of $v_{i}$ over the input feature map $\textbf{x}$; and 2) aggregating the values of neighboring vertexes and $v_{i}$ itself via weighted summation.

\subsection{Central Difference Graph Convolution}
\label{Central Difference Graph Convolution}
Central difference convolution (CDC) introduces the central-oriented local gradient features to enhance model's discrimination capacity, which has been applied in many fields, such as face anti-spoofing\cite{yu2020searching,yu2020fas,yu2020multi,yu2021dual}, remote heart rate measurement\cite{yu2020autohr} and gesture recognition\cite{yu2021searching}, and has achieved great results. Inspired by the CDC networks, we propose to integrate spatial gradient information into the graph convolution operator to enhance its representation and generalization capability.

The proposed central difference graph convolution also consists of two steps, i.e., sampling and aggregation. The sampling step is similar to vanilla graph convolution while the aggregation step is different. As shown in Fig. \ref{CDGC}(a), central difference graph convolution prefers to aggregate the center-oriented gradient of sampled vertexes, which is expressed as:
\begin{equation}
\textbf{y}\left(v_{i}\right)=\sum_{v_{j} \in{R}_{i}} \frac{1}{Z_{i j}} \textbf{w}\left(l_{i}\left(v_{j}\right)\right) \cdot\left(\textbf{x}\left(v_{j}\right)-\textbf{x}\left(v_{i}\right)\right)
\label{central difference}
\end{equation}
When $v_{j}=v_{i}$, the gradient value is always equal to zero relative to the central location $v_{i}$ itself.

For action recognition, existing works focus on establishing dynamic or action specific connections between nodes, so as to aggregate information from either explicitly or implicitly associated nodes. However, the gradient between connected nodes is also potentially useful according to the work of CDC~\cite{yu2020searching}. Hence, the combination of vanilla graph convolution and central difference graph convolution is expected to provide robust and differentiated modeling capability. Thus, the above-mentioned central difference graph convolution is generalized to:
\begin{equation}
\begin{aligned}
\textbf{y}\left(v_{i}\right)&=\alpha \cdot \sum_{v_{j} \in R_{i}} \frac{1}{Z_{i j}} \textbf{w}\left(l_{i}\left(v_{j}\right)\right) \cdot\left(\textbf{x}\left(v_{j}\right)-\textbf{x}\left(v_{i}\right)\right)
\\&+(1-\alpha) \cdot \sum_{v_{j} \in R_{i}} \frac{1}{Z_{i j}} \textbf{w}\left(l_{i}\left(v_{j}\right)\right) \cdot \textbf{x}\left(v_{j}\right) \label{central difference graph convolution}
\end{aligned}
\end{equation}
where hyperparameter $\alpha\in[0,1]$ tradeoffs the contribution between nodes and gradient information. Eq.(\ref{central difference graph convolution}) is referred to as central difference graph convolution (CDGC).

\begin{figure}
	\centering
	\includegraphics[height = 42mm, width = 66mm]{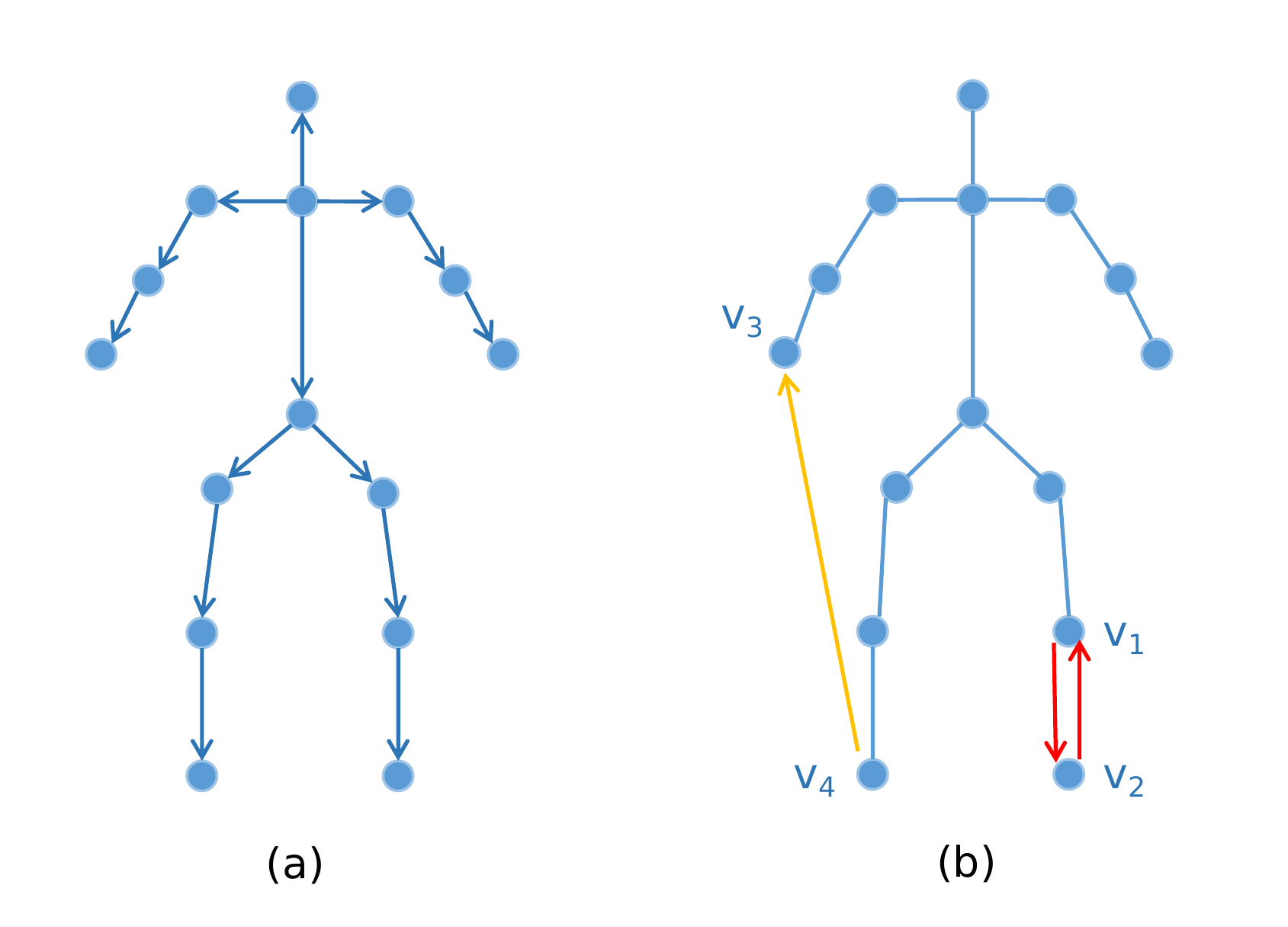}
	\setlength{\abovecaptionskip}{-0.1cm}
	\caption{Illustration of the difference between the gradient information extracted by CDGC and the commonly used bone modality. \textbf{(a)} Illustration of the bone modality with fixed source-target joint directions. The arrows indicate the directions of the bones. \textbf{(b)} Direction representation of gradient information extracted by CDGC. The red arrows indicate that the gradient direction centered at $v_{1}$ points from $v_{1}$ to $v_{2}$, and the direction centered at $v_{2}$ is the opposite. The yellow arrow indicates the direction of the gradient of $v_{3}$ when calculating CDGC with $v_{4}$ as the center, while $v_{4}$  is not directly connected to $v_{3}$.}
\label{difference}
\end{figure}

Notice that the gradient information in CDGC is different from that carried by the commonly used bone modality. First, each bone points from its source joint to the target joint. The source joint is defined as the closer one to the center of gravity of skeleton, and the target joint is defined as the further one. Thus, the direction of each bone is fixed, as shown in Fig. \ref{difference}(a). However, in CDGC, when one of two adjacent nodes is taken as the center node, the direction of the gradient vector is opposite to that obtained when the other node is taken as the center node. According to Eq.(\ref{central difference}) and as shown in Fig. \ref{difference}(b), when $v_{1}$ is the center, the gradient direction of $v_{2}$ relative to $v_{1}$ points from $v_{1}$ to $v_{2}$. When $v_{2}$ is the center, the gradient direction of $v_{1}$ relative to $v_{2}$ points from $v_{2}$ to $v_{1}$, which is opposite to the former. Thus, the information they captured is different. Second, when a connection is established between two nodes that are not connected in the nature skeleton graph, the difference between them is more prominent. If there is a coordination between hand $v_{3}$ and feet $v_{4}$ in an action, the gradient information of $v_{3}$ will be captured by the proposed CDGC when calculating the feature of $v_{4}$. However, there is no bone between the two nodes.

\subsection{Implementation}
\label{Implementation}
A vanilla graph convolution is implemented~\cite{yan2018spatial} as:

\begin{equation}
\textbf{Y}=\textbf{A} \cdot \textbf{X} \cdot \textbf{W}
\end{equation}

Here the weight vectors of multiple output channels are stacked to form the weight matrix $\textbf{W}$. $\textbf{A}$ is a normalized adjacency matrix. In order to simplify the expression of the subsequent formula, we omit the non-linear activation function here. In practice, the input feature map is represented as a tensor of $(C,T,V)$ dimensions, where $C,T$ and $V$denote the number of channels, frames, and vertexes, respectively. The graph convolution is implemented by multiplying the input feature map with the normalized adjacency matrix $\textbf{A}$ on the second dimension and then performing a standard 2D convolution. 

Eq.(\ref{central difference}) can be rewritten to:
\begin{equation}
\begin{aligned}
\textbf{y}\left(v_{i}\right) &=\sum_{v_{j} \in{R}_{i}} \frac{1}{Z_{i j}} \textbf{w}\left(l_{i}\left(v_{j}\right)\right) \cdot \textbf{x}\left(v_{j}\right) \\&- \sum_{v_{j} \in{R}_{i}} \frac{1}{Z_{i j}} \textbf{w}\left(l_{i}\left(v_{j}\right)\right) \cdot \textbf{x}\left(v_{i}\right)
\end{aligned}
\end{equation}

In implementation, the summation of trainable parameters in front of $\textbf{x}(v_{i})$ can be obtained by summing the second dimension of the adjacency matrix. Therefore, the implementation of the central difference graph convolution can be expressed as:
\begin{equation}
\textbf{Y}=\left(\textbf{AX} - \hat{\textbf{A}} \odot \textbf{X}\right) \textbf{W}
\end{equation}
where $\hat{\textbf{A}}$ is obtained by first summing the second dimension of the adjacency matrix $\textbf{A}$, i.e. $\sum_{j} \textbf{A}_{i j}$, and then broadcasting the resulting $ N \times 1 $ vector $\bar{\textbf{A}}$ to $ N \times C$ matrix $\hat{\textbf{A}}$ with $\hat{\textbf{A}} = \bar{\textbf{A}} \cdot \textbf{1} $, with \textbf{1} being the vector of $ 1 \times C $ dimensionality. $\odot$ represents the element-wise Hadamard product. Therefore, the implementation of CDGC is formulated as:

\begin{equation}
\begin{aligned}
\textbf{Y} &=\alpha \cdot (\textbf{AX}-\hat{\textbf{A}} \odot \textbf{X})\textbf{W} + (1-\alpha) \cdot \textbf{AXW} \\& =(\textbf{AX} - \alpha \cdot \hat{\textbf{A}} \odot \textbf{X})\textbf{W}
\end{aligned}
\label{equation7}
\end{equation}

According to Eq.(\ref{equation7}), CDGC can be easily implemented by a few lines of code in PyTorch \cite{paszke2017automatic} or TensorFlow \cite{abadi2016tensorflow}.

\subsection{Accelerated Version of CDGC}
\label{Accelerated Version of CDGC}
In general, implementation of graph convolution depends on an adjacency matrix. The adjacency matrix represents explicit and implicit connections between nodes, then features of the connected nodes are aggregated. However, the GCNs based on an adjacency matrix is usually computationally expensive, and the calculation process is complex and time-consuming. Recently, an advanced graph convolution model called shift-GCN\cite{cheng2020skeleton} is proposed. It utilizes a simple graph shift operation and point-wise convolutions to aggregate neighboring node features instead of regular graph convolutions. This greatly speeds up the training process of GCN. Our accelerated version of CDGC is also based on shift operations, and therefore is also computationally efficient. Moreover, the spatial shift operation makes the receptive field of each node cover the full skeleton graph, and the features of a node after shifting are composed of the features of all other nodes in the graph. Derived from this, we propose a simple solution to realize our CDGC: making difference between the original features and the shifted features.

\begin{figure}
	\centering
	\includegraphics[height = 31mm, width = 86mm]{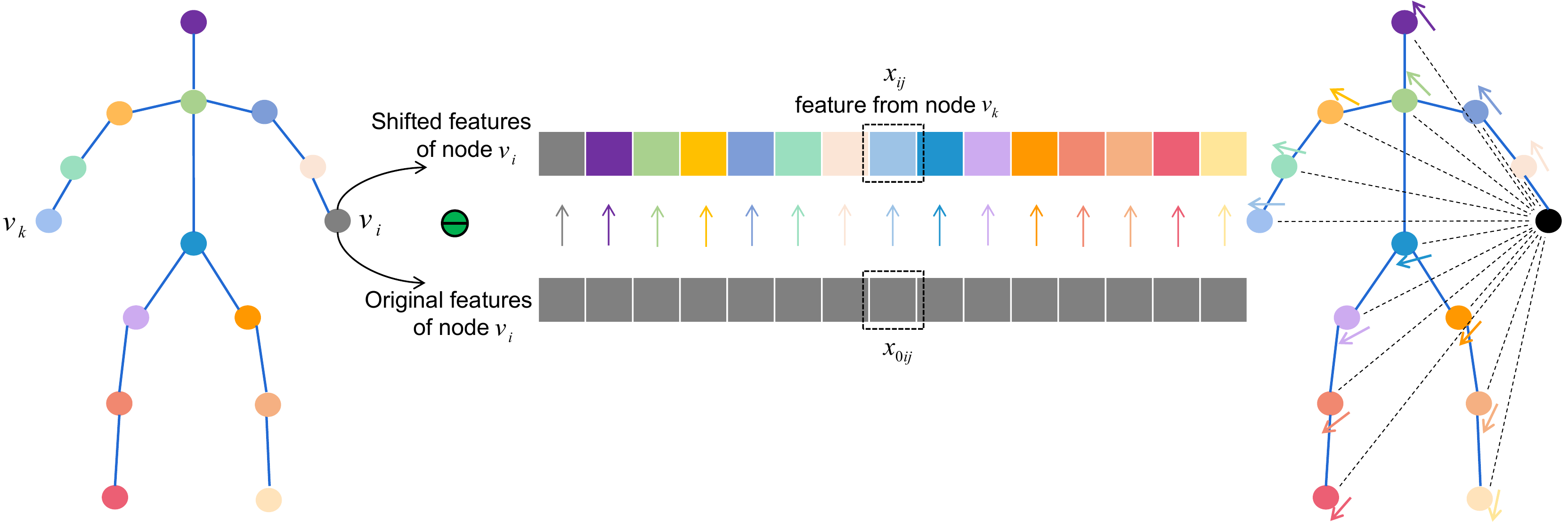}
	\setlength{\abovecaptionskip}{0.1cm}
	\caption{Illustration of accelerated version of CDGC. The gray squares represent the original features of the central node $v_{i}$ (right hand). The colored squares represent its shifted features, with each square feature (channel) having the same color as its corresponding node in the skeleton graph that the feature is from (Please refer to shift-GCN\cite{cheng2020skeleton} for more details). The green minus sign means to subtract the original feature vector from the shifted feature vector. Arrows of different colors indicate the gradient direction of nodes in the human skeleton graph relative to the right hand.}
	\label{Accelerated_CDGC}
\end{figure}

In Fig.\ref{Accelerated_CDGC}, we take a single node $v_{i}$ (right hand) as an example and draw its original feature vector and shifted feature vector. Assuming that $x_{0 i j}$ represents the feature of $j_{t h}$ channel of node $v_{i}$. $x_{i j}$ represents the feature of $j_{t h}$ channel of node $v_{i}$ after shift, which is from one of the channels of node $v_{k}$. Performing $x_{i j}-x_{0 i j}$ obtains the gradient feature of $j_{t h}$ channel for node $v_{k}$ relative to $v_{i}$. By subtracting the original features from the shifted features, the gradient features 
including all body nodes relative to right hand are obtained. In general, given a spatial skeleton feature map $\textbf{X}_{\textbf{0}} \in \mathbb{R}^{N \times C}$, where $N$ stands for the number of joints in skeleton graph, and $C$ is the number of channels of each node. $\textbf{X} \in \mathbb{R}^{N \times C}$ represents the feature map of the shifted spatial skeleton. Performing $\textbf{X}-\textbf{X}_{\textbf{0}}$ can obtain the gradient features of all the remaining nodes in the skeleton graph with respect to their central nodes.

In addition, in the accelerated CDGC, the connection strength between different nodes is the same. But the importance of human skeletons is different. Therefore, we introduce a learnable mask $\textbf{M}$ and multiply it with the obtained feature map to adjust the importance of different connections. The accelerated CDGC can be expressed as:

\begin{equation}
\textbf{Y}=\textbf{W}(\textbf{X}-\textbf{X}_{\textbf{0}})\textbf{M}
\end{equation}
where $\textbf{W}$ is the weight matrix composed of weight vectors of multiple output channels. As described in Sec. \ref{Central Difference Graph Convolution}, both node information captured by vanilla graph convolution and gradient information extracted by CDGC are useful for recognition. Hence, we combine the ordinary shift graph convolution with the accelerated CDGC to construct an effective graph convolution operator. This operator can be expressed as:
\begin{equation}
\begin{aligned}
\textbf{Y} &=\alpha \cdot \textbf{W}(\textbf{X}-\textbf{X}_{\textbf{0}})\textbf{M} + (1-\alpha) \cdot \textbf{WXM} \\& =\textbf{W}(\textbf{X}-\alpha \cdot \textbf{X}_{\textbf{0}})\textbf{M}
\end{aligned}
\label{equation9}
\end{equation}

where $\alpha\in[0,1]$ controls the accelerated CDGC term. The higher value of $\alpha$ is, the more important of central difference gradient information is. The accelerated CDGC is implemented according to Eq.(\ref{equation9}).

\section{Experiments}
\label{Experiments}

The proposed CDGC is evaluated on two popular large-scale skeleton datasets, i.e., NTU RGB+D 60\cite{shahroudy2016ntu} and NTU RGB+D 120\cite{liu2019ntu}. Results are compared with the state-of-the-art methods.

\subsection{Datasets and Evaluation Metrics}

\textbf{NTU RGB+D 60}\cite{shahroudy2016ntu} is currently the most widely used indoor-captured action recognition dataset, which contains a total of 56,880 3D skeleton video clips. The clips are captured from three cameras with different settings. All video clips contain a total of 60 human action classes including both single-actor and two-actor actions. Original work\cite{shahroudy2016ntu} suggests two evaluation protocols: Cross-Subject (CS) and the Cross-View (CV). In the CS evaluation, the dataset is divided into training and testing sets according to the subjects. The training set contains 40,320 videos from 20 subjects, and the rest 20 subjects with 16,560 video clips are used for testing. In the CV evaluation, dataset is divided by the camera ID number. The 37,920 videos captured from camera two and three are used in the training and the 18,960 videos from camera one are used for testing. We report the Top-1 accuracy on both protocols. For inputs with more than one stream, e.g., bones, a score-level fusion result is reported.
 
\textbf{NTU RGB+D 120}\cite{liu2019ntu} is more challenging since it involves more subjects and action categories.  The dataset contains 114,480 action samples in 120 action classes. Samples are captured by 106 volunteers with three cameras views. This dataset contains 32 setups, and each setup denotes a specific location and background. According to\cite{liu2019ntu}, this dataset suggests two evaluation metrics: Cross-Subject and Cross-Setup. The former one splits subjects in half to training and testing parts while the latter one divides the samples based on the camera setup IDs. 
 
\subsection{Experimental Settings}

Our model is based on shift-GCN. Specifically, the model consists of one input block and nine basic blocks. In each basic block, we use the proposed accelerated  CDGC to learn spatial features, and the temporal correlation is still modeled by temporal shift graph convolution. Both of them followed by a batch normalization (BN) layer and a ReLU layer.  A residual connection is added to each basic block to avoid the degradation problem with the increase of network depth, and make the network easy to optimize and fast to converge. A BN layer is added at the beginning to normalize the input graph, and a global average pooling layer is added at the end to pool the feature maps of different samples into the same size. Finally, the output feature is sent to a SoftMax classifier to generate the prediction for action recognition.

All experiments are conducted on the PyTorch deep learning framework \cite{paszke2017automatic}. SGD with Nesterov momentum (0.9) is used to train the model for 140 epochs. Learning rate is set to 0.1 and divided by 10 at epoch 60, 80 and 100.  Cross-entropy is selected as the loss function to back-propagate gradients. For NTU RGB+D 60 and NTU RGB+D 120, the batch size is 48, and we adopt the same data preprocessing as in \cite{shi2019two}. The experiments are performed on two NVIDIA TitanXP GPU.

\begin{figure}
	\centering
	\includegraphics[height = 68mm, width = 54mm]{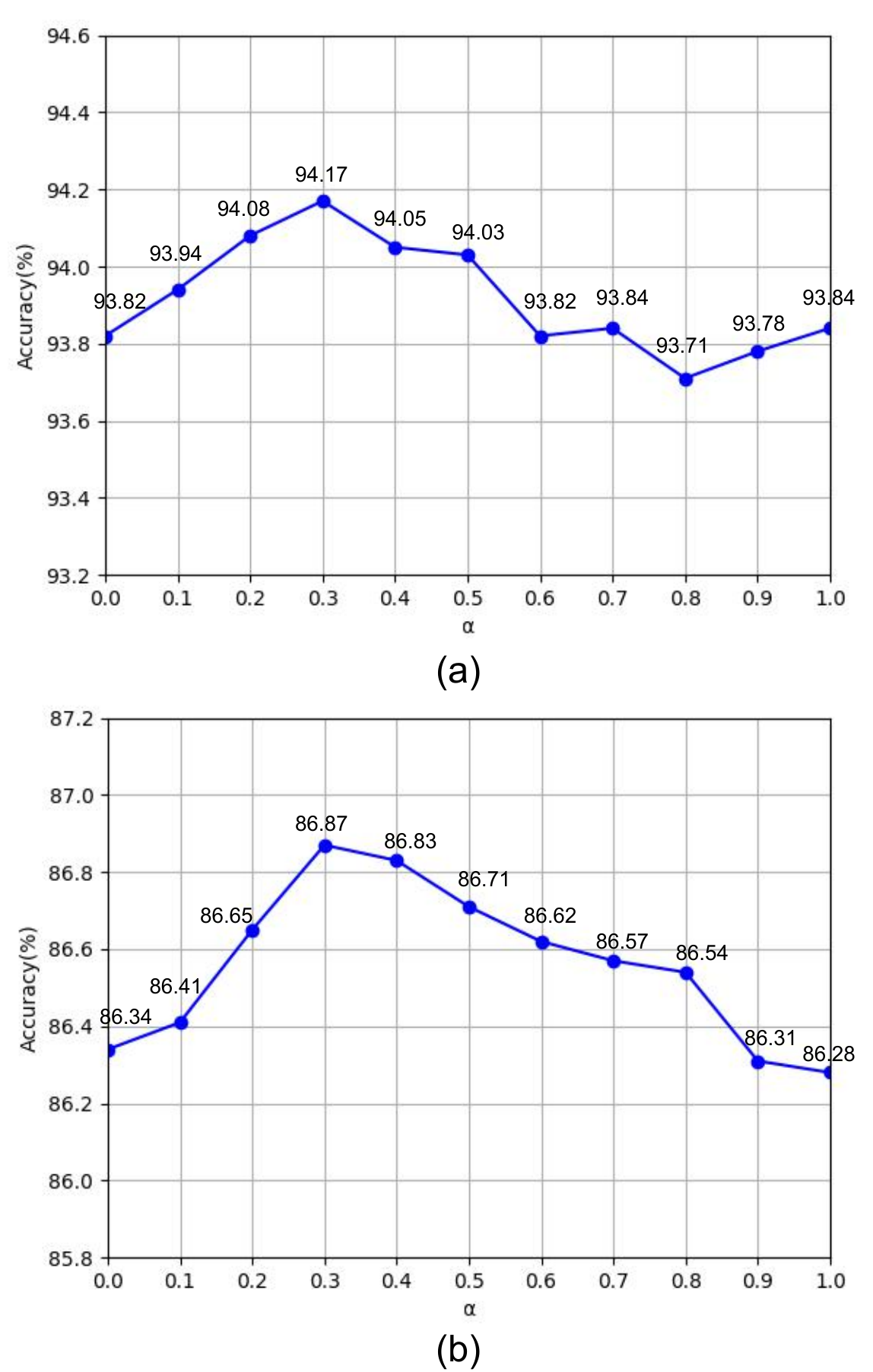}
	\setlength{\abovecaptionskip}{0.cm}
	\caption{Impact of different $\alpha$ on the accuracy of CDGC by following \textbf{(a)} the cross-view protocol, \textbf{(b)} the cross-subject protocol on NTU RGB+D 60 dataset.}
\label{alpha}
\end{figure}

\subsection{Ablation Study}

In this section, we use 2s-AGCN \cite{shi2019two} as the backbone to evaluate the effectiveness of our method. All ablation studies are conducted on the NTU RGB+D 60 dataset.

\subsubsection{Impact of $\alpha$ in CDGC}

In these experiments, we use joint data as input and follow the cross-view and cross-subject protocols, respectively. According to Eq.(\ref{central difference graph convolution}), $\alpha$ controls the contribution of the spatial gradient cues, i.e., the higher $\alpha$, the more gradient information are included. For the cross-view protocol, as shown in Fig.\ref{alpha}(a), when 0 $<\alpha<$ 0.6, the performance of CDGC is better than vanilla graph convolution ($\alpha$ = 0, Accuracy = 93.82$\%$), indicating that the gradient information captured by CDGC improves recognition. The best performance is obtained when $\alpha$ = 0.3 (Accuracy = 94.17$\%$). In the cross-subject protocol, as shown in Fig.\ref{alpha}(b), the best accuracy is also obtained at $\alpha$ = 0.3 (Accuracy = 86.87$\%$). Thus, we use this optimal value ($\alpha$ = 0.3) in the following experiments.

\subsubsection{Impact of CDGC}

In order to verify the capability of CDGC, we replace the spatial GCN module in 2s-AGCN with the CDGC to test whether the performance will drop or increase.

\begin{table}
\centering
\caption{Ablation study about CDGC on NTU RGB+D 60 dataset.}
\begin{tabular}{|c|c|c|c|}
\hline
              & data       & X-sub        & X-view     \\
\hline
2s-AGCN       & Joint      &   86.3$\%$   & 93.8$\%$   \\
2s-AGCN(CDGC) & Joint      &   86.9$\%$   & 94.2$\%$   \\
\hline
2s-AGCN       & Bone       &   87.0$\%$   & 93.2$\%$   \\
2s-AGCN(CDGC) & Bone       &   87.5$\%$   & 93.9$\%$   \\
\hline
2s-AGCN       & 2s         &  88.5$\%$    & 95.1$\%$   \\
2s-AGCN(CDGC) & 2s         &  89.1$\%$    & 95.5$\%$   \\
\hline
\end{tabular}
\label{Impact of CDGC}
\end{table}

\begin{figure}
	\centering
	\includegraphics[height = 34mm, width = 88mm]{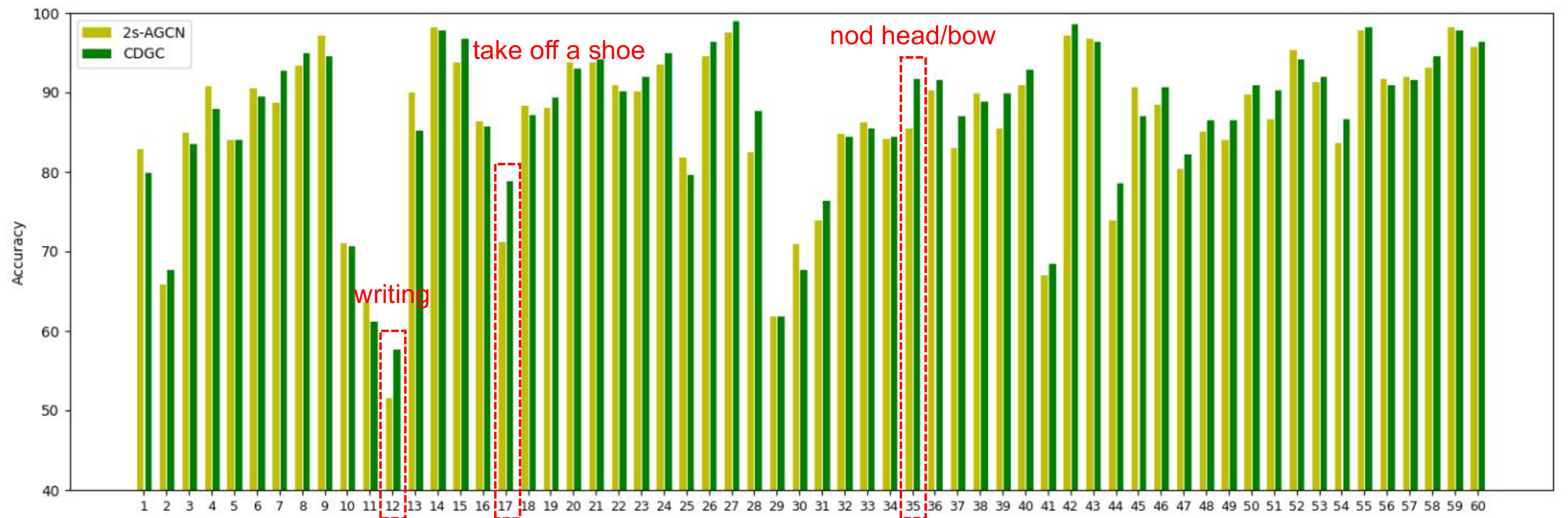}
	\setlength{\abovecaptionskip}{-0.3cm}
	\caption{Accuracy comparison of each action class on the NTU RGB+D 60 dataset between 2s-AGCN and CDGC.}
	\label{accuracy_class}
\end{figure}

The results are shown in Table \ref{Impact of CDGC}. The recognition results of the 2s-AGCN model are run by ourselves. As seen, under the cross-subject evaluation, the accuracy is improved by 0.6 and 0.5 percentage points on the joint stream and bone stream, respectively. After fusion of them, the recognition performance is improved by 0.6 percentage points. Under the cross-view evaluation, we improve the accuracy by 0.4 and 0.7 percentage points on the joint stream and bone stream, respectively. After fusion, the recognition accuracy is improved by 0.4 percentage points. Also, Fig.\ref{accuracy_class} shows the accuracy comparison of each action class between 2s-AGCN and CDGC. It can be seen that the recognition accuracy of our CDGC in categories "writing", "take off a shoe" and "nod head/bow" is significantly higher than that of the baseline. For the three classes, the motion directions of the related collaborative joints and the relative motion relationship among them are important. The CDGC can not only aggregate features of neighboring nodes, but also capture the differential gradient information among them, which can reflect the motion direction and amplitude of adjacent nodes, so as to improve recognition. In addition, it should be noted that CDGC does not increase the number of parameters, but it has performance advantage over the vanilla graph convolution.

\subsubsection{Impact of accelerated CDGC on network training}

\begin{table}
\centering
\caption{Ablation study on NTU RGB+D 60 dataset about contribution of accelerated CDGC to the network training.}
\begin{tabular}{|c|c|c|c|}
\hline
   & Time Consumption   & Params   & Convergence  \\
\hline
CDGC              & 53min   & 3.47M    &30th epoch   \\
\hline
Accelerated CDGC  & 10min   & 0.69M    &60th epoch   \\
\hline
\end{tabular}
\label{network training}
\end{table}

\begin{figure}
	\centering
	\includegraphics[height = 38mm, width = 62mm]{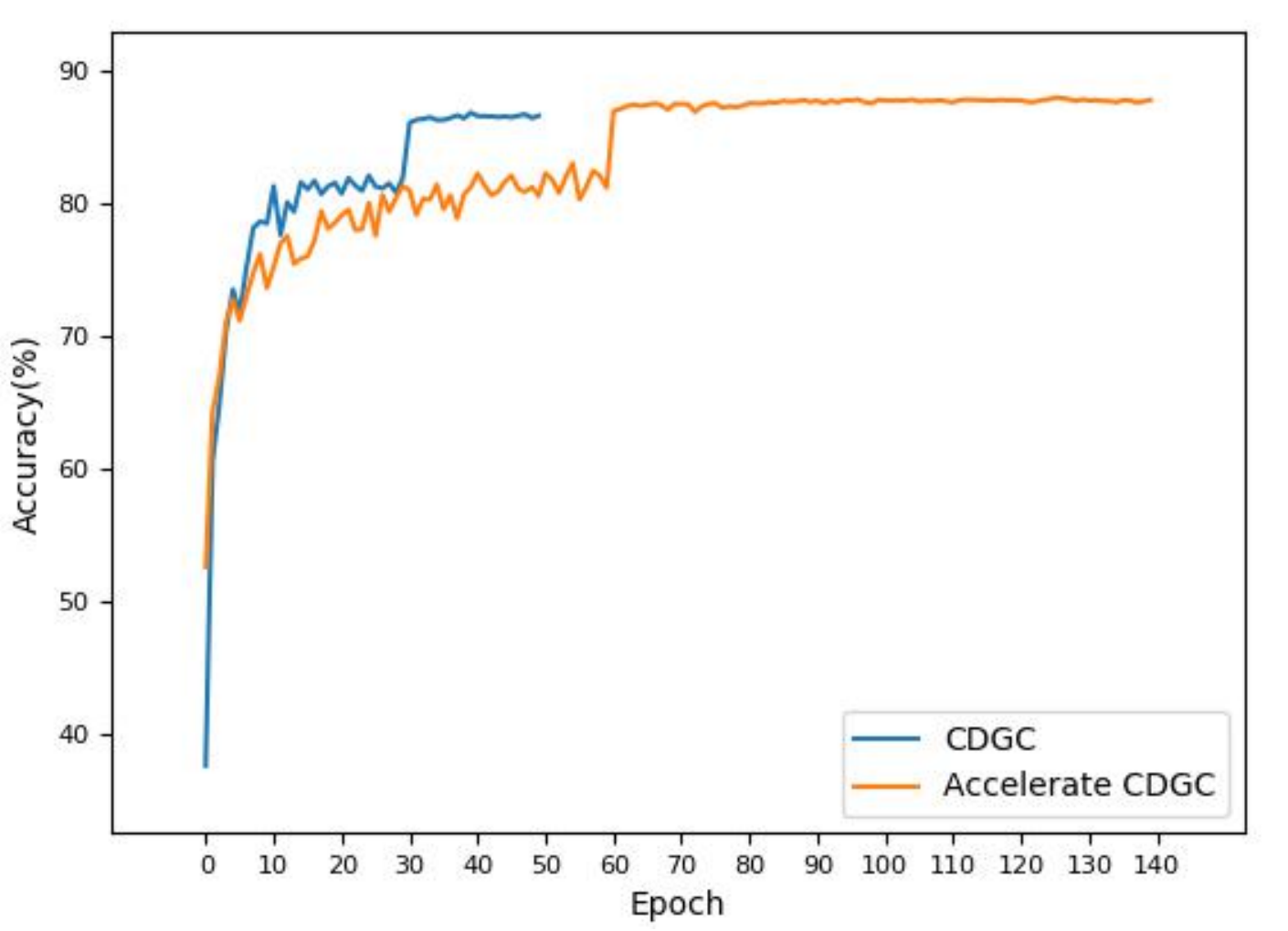}
	\setlength{\abovecaptionskip}{0.cm}
	\caption{Training accuracy changing curves of the CDGC network and the accelerated version of CDGC.}
	\label{accuracy_curve}
\end{figure}

In order to verify the contribution of accelerated CDGC to the network training compared to the ordinary CDGC in Sec.\ref{Central Difference Graph Convolution}. We compare the two in three aspects: time consumption per epoch, model parameters and number of epochs to converge. As shown in Table \ref{network training}, the accelerated CDGC is significantly better than the ordinary one in the first two items as expected. Fig.~\ref{accuracy_curve} shows the training accuracy. We can see that CDGC converges faster and needs less number of training epochs than the accelerated CDGC. Table \ref{CDGC and Accelerated CDGC} compares the recognition accuracy of the two versions of CDGC. It can be seen that the recognition accuracy of the accelerated CDGC is higher than the original one for all the stream settings. This is mainly attributed to that the accelerated CDGC makes the receptive field of each node cover all the nodes in the skeleton graph.

\begin{table}[!th]
\centering
\caption{Comparison of recognition accuracy($\%$) between CDGC and the accelerated CDGC on the NTU RGB+D 60 dataset.}
\resizebox{.95\columnwidth}{!}{
\begin{tabular}{|c|c|c|c|c|c|c|}
\hline
\multirow{2}{*}{NTU RGB+D 60} & \multicolumn{3}{c|}{X-sub} & \multicolumn{3}{c|}{X-view} \\
\cline{2-7}
& Joint & Bone & 2s & Joint & Bone & 2s  \\
\hline
CDGC             & 86.9  & 87.5  & 89.1  & 94.2  & 93.9  & 95.5   \\
\hline
Accelerated CDGC & 88.0  & 88.1  & 89.7  & 94.8  & 95.0  & 95.9   \\
\hline
\end{tabular}}
\label{CDGC and Accelerated CDGC}
\end{table}

\subsection{Comparison with the state-of-the-art methods}

In this experiment, we combine the accelerated CDGC with shift-GCN and compare it with the state-of-the-art methods on NTU RGB+D 60 and  NTU RGB+D 120. In addition, as the key hyperparameter of CDGC, $\alpha$ in each layer of the network can also be learned together with the network training. Thus, we also conduct experimental exploration with the learnable $\alpha$ using the same experimental settings as the chosen optimal value 0.3. Like \cite{cheng2020skeleton}, we build four stream networks: joint stream, bone stream, joint motion stream and bone motion stream. For comparison, we report the results of single-stream(joint stream), double-stream(joint stream and bone stream) and four-stream(all). In the Tables \ref{SOTA60} and \ref{SOTA120}, we abbreviate them as 1s, 2s and 4s, respectively.

\begin{table}[!th]
\centering
\caption{Comparison of recognition accuracy($\%$) with state-of-the-art methods on NTU RGB+D 60 dataset. 1s, 2s and 4s denote single-stream, double-stream and four-stream, respectively.}
\begin{tabular}{|c|c|c|}
\hline
methods             & X-sub     & X-view  \\
\hline
TCN\cite{kim2017interpretable}          & 74.3      & 83.1     \\
VA-LSTM\cite{zhang2017view}             & 79.2      & 87.7     \\
ST-GCN\cite{yan2018spatial}             & 81.5      & 88.3     \\
DPRL\cite{tang2018deep}                 & 83.5      & 89.8      \\
SR-TSL\cite{si2018skeleton}             & 84.8      & 92.4      \\
STGR-GCN\cite{li2019spatio}             & 86.9      & 92.3      \\
GR-GCN\cite{gao2019optimized}           & 87.5      & 94.3     \\
SAR-NAS\cite{zhang2020sar}              & 86.4      & 94.3      \\
2s AS-GCN\cite{li2019actional}          & 86.8      & 94.2      \\
2s-AGCN\cite{shi2019two}                & 88.5      & 95.1      \\
4s Directed-GNN\cite{shi2019skeleton}   & 89.9    & 96.1       \\
1s Shift-GCN\cite{cheng2020skeleton}    & 87.8      & 95.1     \\
2s Shift-GCN\cite{cheng2020skeleton}    & 89.7      & 96.0     \\
4s Shift-GCN\cite{cheng2020skeleton}    & 90.7      & 96.5     \\
\hline
ours($\alpha$=0.3 1s)       & \textbf{88.0}      & 94.8     \\
ours($\alpha$=0.3 2s)       & 89.7               & 95.9     \\
ours($\alpha$=0.3 4s)       & \textbf{91.0}      & 96.4     \\
\hline
ours(learnable $\alpha$ 1s)   & \textbf{88.0}      & 94.9     \\
ours(learnable $\alpha$ 2s)   & 89.7               & 95.8     \\
ours(learnable $\alpha$ 4s)   & \textbf{90.9}      & 96.5     \\
\hline
\end{tabular}
\label{SOTA60}
\end{table}

\begin{table}
\centering
\caption{Comparison of recognition accuracy($\%$) with state-of-the-art methods on NTU RGB+D 120 dataset. 1s, 2s and 4s denote single-stream, double-stream and four-stream, respectively.}
\begin{tabular}{|c|c|c|}
\hline
methods                 & X-sub     & X-setup  \\
\hline
Dynamic Skeleton\cite{hu2015jointly}          & 50.8      &  54.7    \\
FSNet\cite{liu2019skeleton}                   & 59.9      & 62.4     \\
MT-CNN-RotClips\cite{ke2018learning}          & 62.2      & 61.8     \\
Pose Evolution Map\cite{liu2018recognizing}   & 64.6      & 66.9     \\
ST-GCN\cite{yan2018spatial}                   & 72.4      & 71.3     \\
AS-GCN\cite{li2019actional}                   & 77.7      & 78.9      \\
1s Shift-GCN\cite{cheng2020skeleton}          & 80.9      & 83.2     \\
2s Shift-GCN\cite{cheng2020skeleton}          & 85.3      & 86.6     \\
4s Shift-GCN\cite{cheng2020skeleton}          & 85.9      & 87.6     \\
\hline
ours($\alpha$=0.3 1s)       & \textbf{81.3}      & 83.2            \\
ours($\alpha$=0.3 2s)       & 85.0               & \textbf{86.7}    \\
ours($\alpha$=0.3 4s)       & \textbf{86.3}      & \textbf{87.8}    \\
\hline
ours(learnable $\alpha$ 1s)       & \textbf{81.1}      & \textbf{83.3}   \\
ours(learnable $\alpha$ 2s)       & 85.0               & \textbf{86.7}    \\
ours(learnable $\alpha$ 4s)       & \textbf{86.3}      & \textbf{87.8}    \\
\hline
\end{tabular}
\label{SOTA120}
\end{table}

The bold numbers indicate that our method has improved the accuracy of skeleton-based action recognition compared with the corresponding baseline. It can be seen from Table \ref{SOTA60} that our model achieves the best performance for the cross-subject protocol, i.e. 91.0$\%$, 0.3 percentage points higher than the baseline shift-GCN with optimal value $\alpha$ = 0.3. NTU RGB+D 120 dataset is more challenging than NTU RGB+D 60. We can see from Table~\ref{SOTA120} that the performance of our model is 0.4 and 0.2 percentage points higher than the baseline for both cross-subject and cross-setup protocols, respectively, and has achieved the best results among the reported to date. We can also see from the two tables that the accuracy of learnable $\alpha$ is almost the same as that of $\alpha$ = 0.3, which proves that $\alpha$ = 0.3 is an optimal choice. Note that the number of parameters in our model is equal to the baseline. The results have demonstrated the effectiveness of CDGC.

\section{Conclusion}
\label{sec:conclusion}
In this paper, we propose a new graph convolution operator called CDGC, which not only aggregates node information as a vanilla graph convolution operation does but also aggregates gradient information between nodes. Without introducing any additional parameters, CDGC can replace vanilla graph convolution in any existing GCNs. In addition, we have developed an accelerated version of CDGC that greatly improves the training speed of the network. Experiments on two popular large-scale datasets, NTU RGB+D 60 and NTU RGB+D 120, have demonstrated the efficacy of the proposed CDGC. Importantly, the results have verified our hypothesis that {\it improving the graph convolution operator itself could be as effective as improving the topology of a skeleton graph}.



\bibliography{tcsvt_manuscript}

\end{document}